\documentclass[conference]{IEEEtran}
\usepackage{multirow, booktabs, url, makecell, xcolor, tabularx,graphicx}
\graphicspath{{figures/}}
\usepackage{amssymb, amsmath}
\usepackage{colortbl}
\definecolor{lightgray}{gray}{0.9}
\usepackage{array}
\newcolumntype{C}[1]{>{\centering\let\newline\\\arraybackslash\hspace{0pt}}m{#1}}
\newcolumntype{L}[1]{>{\raggedright\let\newline\\\arraybackslash\hspace{0pt}}m{#1}}
\newcolumntype{R}[1]{>{\raggedleft\let\newline\\\arraybackslash\hspace{0pt}}m{#1}}
\title{Zero-shot Bilingual App Reviews Mining with Large Language Models}
\author{
Jialiang Wei$^*$, Anne-Lise Courbis$^*$, Thomas Lambolais$^*$, \\Binbin Xu$^*$, Pierre Louis Bernard$^{**}$ and Gérard Dray$^*$
\\[6pt]
$^{*}$: {\normalsize EuroMov Digital Health in Motion, Univ Montpellier, IMT Mines Ales, Ales, France}\\
$^{**}$: {\normalsize EuroMov Digital Health in Motion, Univ Montpellier, IMT Mines Ales, Montpellier, France}\\
\texttt{$^{*}$: {\small firstname.lastname@mines-ales.fr} $^{**}$: {\small firstname.lastname@umontpellier.fr}}
}

\usepackage{tcolorbox}
\begin{document}
\pagestyle{plain} %
\maketitle

\begin{abstract}
App reviews from app stores are crucial for improving software requirements. 
A large number of valuable reviews are continually being posted, describing software problems and expected features.
Effectively utilizing user reviews necessitates the extraction of relevant information, as well as their subsequent summarization.
Due to the substantial volume of user reviews, manual analysis is arduous. 
Various approaches based on natural language processing (NLP) have been proposed for automatic user review mining.
However, the majority of them requires a manually crafted dataset to train their models, which limits their usage in real-world scenarios.

In this work, we propose Mini-BAR, a tool that integrates large language models (LLMs) to perform zero-shot mining of user reviews in both English and French.
Specifically, Mini-BAR is designed to
(i) classify the user reviews, 
(ii) cluster similar reviews together, 
(iii) generate an abstractive summary for each cluster and 
(iv) rank the user review clusters. 
To evaluate the performance of Mini-BAR, we created a dataset containing 6,000 English and 6,000 French annotated user reviews and conducted extensive experiments.
Preliminary results demonstrate the effectiveness and efficiency of Mini-BAR in requirement engineering by analyzing bilingual app reviews.
\end{abstract}

\section{Introduction}

App stores, such as Google Play and Apple App Store, allow users to express their feedback on downloaded apps. 
This feedback is in form of rating scores and text reviews. 
The latter contains praise and dispraise, user experience, problem reports, and feature requests~\cite{Pagano2013}. 
App reviews are important for app success. As evidenced in prior literature, high-user rating scores have positive effects on apps’ sustainability~\cite{Lee2014}. 
Consequently, the design and development teams should take the users' feedback into consideration during the evolution of their application.

Due to the large amount and the redundancy of app reviews, manual analysis is laborious. 
Various approaches based on natural language processing (NLP) have been proposed to reduce the efforts in analyzing user feedback, including the classification, clustering and summarization.
Classification models are commonly employed in the first approach to categorize app reviews into predefined groups, such as feature requests and problem reports~\cite{Chen2014,Maalej2016,Mekala2021,Henao2021,Zhang2023}.
However, even after classification, the volume of reviews within each category remains substantial, making direct analysis impractical.
To tackle this issue, some researchers have proposed grouping reviews that pertain to the same topic~\cite{Scalabrino2019,Stanik2021,Devine2022,Wang2022}. 
As the obtained clusters still contain a relatively large number of user reviews, the manual analysis of each cluster continues to be time-consuming.
Certain techniques attempt to overcome this challenge by selecting the most representative phrases or sentences as summaries for groups of app reviews~\cite{Stanik2021,Devine2022,Sorbo2016,Alshangiti2022,Gao2022}. 
Nevertheless, this extractive summarization approach may not capture all the crucial information present within a given group.
Moreover, existing approaches in user review analysis mainly focus on the English language, with few works on analyzing reviews in other languages~\cite{Stanik2019,Wei2022}.
Furthermore, most existing approaches requires manually crafted dataset for training their classification models. 
The creation of dataset is costly and time consuming, which limits their usage in real-word scenarios.
The objective of this article is therefore to address these gaps with large language models.

Pre-trained language models (PTMs) are deep neural networks previously trained on a vast corpus.
Researchers have observed that large-sized PTMs display different behaviors from smaller PTMs and show surprising abilities (called emergent abilities) in solving a series of complex tasks.
Thus, the research community coins the term ``large language models (LLMs)'' for these large-sized PTMs~\cite{Zhao2023a}.
A remarkable application of LLMs is ChatGPT\footnote{https://openai.com/blog/chatgpt/}, it is fine-tuned from the GPT-3.5~\cite{Brown2020} using Reinforcement Learning from Human Feedback (RLHF), which optimizes the model by interacting with human and learning from human preference.
The Guanaco model is an open-source, finely-tuned LLM, derived through the application of QLoRa's 4-bit tuning approach~\cite{Dettmers2023} on LLaMA base models~\cite{touvron2023llama}.
QLoRA is an efficient fine-tuning approach that reduces memory usage.
Guanaco is available in various parameter sizes, including 7B, 13B, 33B and 65B.

\begin{figure}[!htb]    
    \centerline{\includegraphics[width=0.5\textwidth]{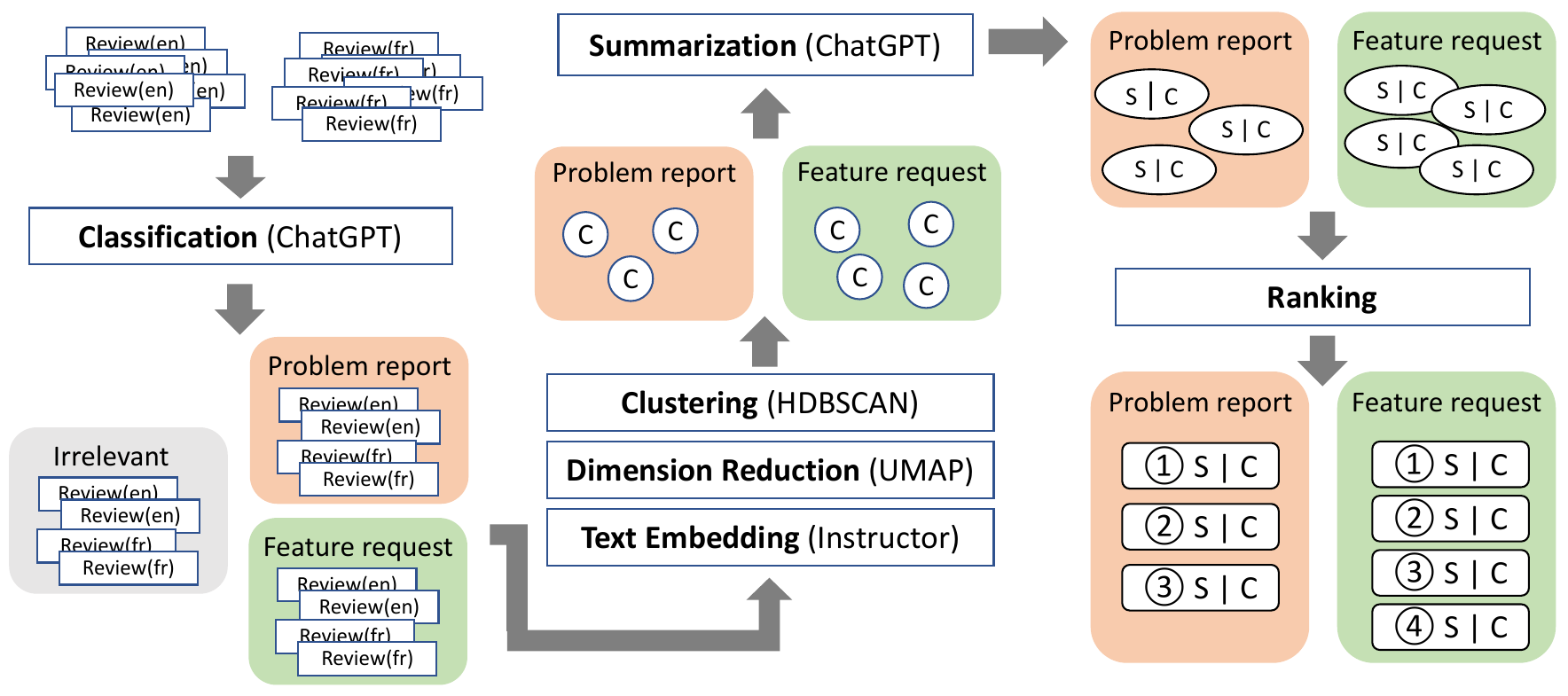}}
    \caption{Overview of Mini-BAR}
    \label{fig:overview}
\end{figure}

In this paper, we propose Mini-BAR, a bilingual approach based on LLMs to:
(i) classify the user reviews into three categories: \textit{feature request}, \textit{problem report} and \textit{irrelevant};
(ii) cluster similar reviews for \textit{feature request} and \textit{problem report}; 
(iii) generate a summary for each cluster of user reviews; 
and (iv) rank the user review clusters.
Figure \ref{fig:overview} depicts an overview of the workflow of Mini-BAR.
We use the same pipeline to process the bilingual app reviews, eliminating the necessity of deploying separate models for each language.
By combining these functionalities, Mini-BAR provides a comprehensive approach for analyzing bilingual app reviews, which can yield valuable insights for app developers and marketers.
We validate the key steps of Mini-BAR by conducting an extensive set of experiments on 12000 annotated user reviews from three Health \& Fitness apps.
The results indicate that Mini-BAR has a satisfactory performance in both classification and clustering tasks, and produced high-quality summaries.
We provide a replication package\footnote{https://github.com/Jl-wei/mini-bar} containing the code, dataset, and experiment setups.

\section{Approach}\label{sec:approach}
Mini-BAR provides support to developers for the analysis of mobile app user reviews through a four-step process.
First, it applies a pre-trained classifier to categorize the user reviews (Section \ref{sec:approach-classification}). 
The second step clusters the user reviews based on their semantic similarity (Section \ref{sec:approach-clustering}).
The third step summarizes the user reviews belonging to the same cluster (Section \ref{sec:approach-summarization}).
The last step is to determine the importance of user review clusters and rank them accordingly (Section \ref{sec:approach-ranking}).
In the following subsections, we will detail each step of Mini-BAR.

\subsection{Classification}\label{sec:approach-classification}
The objective of this step is to automatically classify English and French user reviews into three categories: (F) \textit{feature request}, (B) \textit{problem report}, and (I) \textit{irrelevant}.
A user review may belong to either one of the three categories or both \textit{feature request} and \textit{problem report}.
A user review is considered as \textit{problem report} if it mentions the issues the users have experienced while using the app (\textit{e.g.}, ``Can't sync sleep data since last update").
\textit{Feature requests} reflect users' needs for new functions, new content, or improvements (\textit{e.g.}, ``Please bring a feature to add some custom watch faces \ldots'').
All the other user reviews are \textit{irrelevant} (\textit{e.g.}, ``Best app ever!'').
Classifying the reviews can aid to redirect them to the appropriate software project members. 
For instance, \textit{feature requests} can be delivered to requirements analysts, while \textit{problem reports} can be directed to developers and testers~\cite{Maalej2016}.

The classifier of Mini-BAR is based on ChatGPT, the model we use is \textit{gpt-3.5-turbo}\footnote{https://platform.openai.com/docs/models/gpt-3-5}.
We use the following prompt to classify the app reviews.
\begin{tcolorbox}[fontupper=\small]
\texttt{
Classify the following \{lang\} app review into problem report, feature request or irrelevant. Be concise.\\
```\\
\{review\}\\
```
}
\end{tcolorbox}
Given a user review, its language is detected automatically with Lingua\footnote{https://github.com/pemistahl/lingua-py}, which is an accurate language detector.
The detected language, which could be English, French among others, replaces the $\{lang\}$ variable in the prompt.
And the $\{review\}$ in the prompt is replaced by the user review.
The response of ChatGPT contains a single-phrase label name.
We parse the response with regular expressions to automatically obtain the predicted labels.

\subsection{Clustering}\label{sec:approach-clustering}
The objective of this step is to group English and French user reviews based on their semantic similarity, ensuring that reviews within a group are related to the same topic. 
Through clustering analysis, texts are divided into clusters such that those within a cluster exhibit semantic similarity. 
Currently, the RE community predominantly focuses on clustering English user reviews, leaving little attention to non-English user reviews. 
To address this gap, we propose a bilingual clustering approach that allows the creation of clusters comprising reviews from different languages that share common topics.

\begin{figure}[!htbp]
\centerline{
    \includegraphics[width=0.45\textwidth]{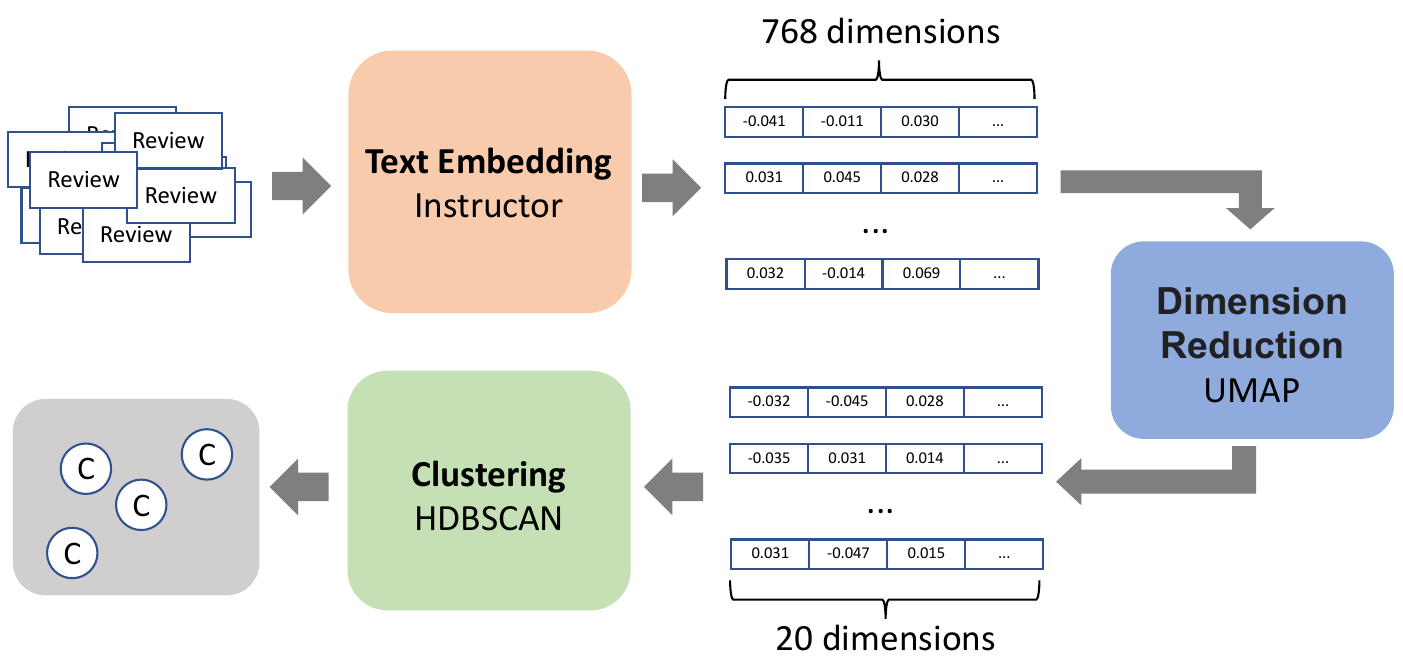}}
    \caption{Overview of clustering}
    \label{fig:clustering}
\end{figure}

In this step, we perform bilingual clustering analysis on user reviews that belong to the same category, namely \textit{feature request} or \textit{problem report}, which were identified in the previous step.
It is worth noting that in the clustering process, the categories of \textit{feature requests} and \textit{problem reports} are processed separately, in order to obtain distinct clusters for each. 
Conversely, user reviews labeled as \textit{irrelevant} are excluded from the clustering analysis.

The user reviews cannot be directly used as input for the clustering algorithm, as they are in a textual format.
Therefore, it is necessary to convert them into embeddings --- numerical vectors in a high-dimensional space --- to enable effective processing.
In this space, similar inputs in different languages are mapped close together.
For example, the embeddings of ``Problème de serveur récurrent" and ``Connection issues to the main server" are in proximity to each other.
We used \textit{Instructor}~\cite{Su2022} to embed English and French app reviews due to its high performance proven in Section \ref{sec:clustering-experiment}. 
\textit{Instructor} is an instruction-finetuned text embedding model that can generate text embeddings tailored to any task (e.g. clustering) by simply providing the task instruction, without any finetuning.
In our case, we have utilized the instruction "Represent the app user review for clustering".
This model generates 768-dimensional embeddings for each app review.

The high dimension of embeddings causes high computation costs. 
Dimension reduction techniques can transforms data from a high-dimensional space into a low-dimensional space and keeps meaningful information of the original data.
As in Stanik et al.~\cite{Stanik2021}, we reduced the embeddings' dimension with Uniform Manifold Approximation and Projection (UMAP). 
The implementation of Mini-BAR utilizes the UMAP Python package\footnote{https://github.com/lmcinnes/umap}, the UMAP parameters are as follows: output dimensionality of $20$, number of neighbors set to $100$, and minimum distance of $0$.

The reduced embeddings of the \textit{feature requests} and \textit{problem reports} are then clustered by HDBSCAN~\cite{McInnes2017}, which has been proven efficient by Devine et al.~\cite{Devine2022} and Stanik et al.~\cite{Stanik2021}. 
We use the HDBCAN\footnote{https://github.com/scikit-learn-contrib/hdbscan} Python package for the implementation of Mini-BAR.
The HDBSCAN parameters include a minimum cluster size of 5.
The minimum cluster size is the smallest grouping size considered as a cluster.
In our study, we chose a minimum cluster size of 5, as we were interested in identifying problems or features that were reported by at least 5 users.

\subsection{Summarization}\label{sec:approach-summarization}
Given the potential magnitude of reviews within clusters, the process of summarization becomes imperative, enabling developers to efficiently grasp the cluster's contents without the need to peruse every individual review.
Large language models (LLMs), such as ChatGPT, has achieved a state-of-the-art performance in cross-lingual summarization \cite{Wang2023}.
In this step, we utilized ChatGPT (\textit{gpt-3.5-turbo}) to generate abstractive English summaries for clusters containing bilingual user reviews.
Our evaluation in Section \ref{sec:evaluation-summarization} proves that ChatGPT outperforms extractive summarization method, and is able to generate high-quality summaries.

To generate the summaries, we utilized the following prompt within ChatGPT. 
The $\{reviews \: list\}$ is replace by a list of user reviews separated by new line break. 
The response generated by ChatGPT represents the summary for that list of user reviews.
Table \ref{tab:cluster_summary_example} presents an example of a manual created cluster and its generated summary.
\begin{tcolorbox}[fontupper=\small]
\texttt{
Please summarize all following app reviews into one English sentence: \\
```\\
\{reviews list\}\\
```
}
\end{tcolorbox}

\begin{figure}[!htb]
\centerline{
    \includegraphics[width=0.3\textwidth]{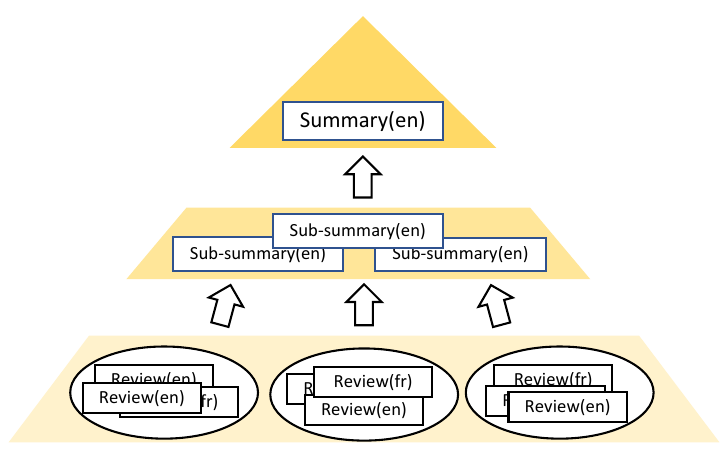}}
    \caption{Overview of summarization for large clusters}
    \label{fig:summarization}
\end{figure}

However, in case the user review clusters contain a large number of reviews and exceed the input length limitation of ChatGPT, it will issue a warning message indicating that ``The message you submitted was too long, please reload the conversation and submit something shorter".
To address this issue, as illustrated in Figure \ref{fig:summarization}, we adopted a hierarchical summarization approach consisting of the following steps:
(i) dividing the reviews belonging to one cluster into multiple groups, each with a maximum of 4000 tokens\footnote{https://platform.openai.com/docs/guides/gpt/chat-completions-vs-completions},
(ii) generating a sub-summary for each group of user reviews, and
(iii) obtaining the final summary by summarizing the sub-summaries.
Since the length of all sub-summaries may also exceed the input limit of ChatGPT, we used a recursive procedure for steps (i) and (ii) by replacing the user reviews with sub-summaries.

\begin{table}[!htb]
\centering
 \caption{Example of a user review cluster and its generated summary}
 \begin{tabular}{p{.45\textwidth}} 
 \toprule
 \multicolumn{1}{c}{\textbf{User reviews}:} \\
- Dommage que la connexion 4g soit indispensable pour fonctionner. \\
- Please for god sake make it to work offline also. \\
- Is not work offline \\
- It used to work offline. Now I have to log in just to see my old data. \\
- Useless without internet. \\
 \midrule
 \multicolumn{1}{c}{\textbf{Summary}:} \\
Users are disappointed that the app requires an internet connection to function and wish it could work offline like it used to. \\
 \bottomrule
 \end{tabular}
\label{tab:cluster_summary_example}
\end{table}

\subsection{Ranking}\label{sec:approach-ranking}
Given the clusters with summaries, the next step is to rank them by their importance.
The goal of this step is to aid in the release planning of app developers.
The importance of a cluster ($\mathit{ClusterScore}$) is determined based on the following characters:
\begin{itemize}
\item The number of reviews present within the cluster ($|\mathit{reviews}|$). 
problems or feature requests reported by more users should be given higher priority compared to those reported by fewer users.
\item The average rating of the cluster ($\overline{\mathit{rating}}$). 
Clusters with lower average ratings should be given higher priority, as they may indicate users' greater dissatisfaction with specific aspects of the app.
\item The number of ``thumbs up" inside the cluster ($|\mathit{thumbs up}|$).
Users on Google Play have the option to click the ``thumbs up" button on reviews that they find helpful.
We posit that the number of ``thumbs up" and the importance of a cluster are positively correlated as the review liked by more users should have a higher priority.
\end{itemize}
Given the weight of $w_{rev}$, $w_{th}$ and $w_{ra}$, which are assigned default values of $1$, $0.1$, and $1$, respectively, the calculation of $\mathit{ClusterScore}$ is defined as follows:
\begin{align}
\mathit{ClusterScore} &= \frac{w_\mathit{rev}\cdot|\mathit{reviews}| + w_{th}\cdot|\mathit{thumbs up}|}{w_\mathit{ra}\cdot\overline{\mathit{rating}}}
\end{align}
The clusters are ranked in decreasing order of $\mathit{ClusterScore}$.

\section{Empirical Evaluation}\label{sec:evaluation}
The objective of this study is to assess the performance of Mini-BAR with respect to three criteria: 
(i) its accuracy in classifying user reviews into one of three predefined categories, namely \textit{feature request}, \textit{problem report}, and \textit{irrelevant};
(ii) its ability to cluster related user reviews that fall into the same topic;
(iii) its ability to provide high-quality summaries of user reviews clusters.
To achieve this goal, we evaluated Mini-BAR's performance on a dataset of 6000 English and 6000 French reviews from three health-related mobile apps.

\subsection{Evaluation of Classification}\label{sec:evaluation-classification}
In this section, we aim to answer the following research question (RQ$_1$):
\textit{How accurate is Mini-BAR in classifying bilingual user reviews ?}

\subsubsection{Dataset}\label{sec:classification-dataset}
The training and evaluation of the classifier require a large number of labeled user reviews.
We relabelled the $6000$ French reviews of three applications (Garmin Connect, Huawei Health and Samsung Health) on Google Play from our previous work \cite{Wei2022}.
Besides, we collected $365,967$ English user reviews from these three applications.
For each application, $2000$ English are randomly sampled for annotation.
In this work, we have labeled $6000$ English reviews and $6000$ French reviews.

We used Prodigy\footnote{https://prodi.gy/} from spaCy to annotate the user reviews. 
We created an annotation guide to clarify the definition of \textit{feature request}, \textit{problem report}, and \textit{irrelevant}.
Four authors of this paper annotated the sampled user reviews and they are finally reviewed by the first author of this paper.
Table \ref{tab:classification-dataset} shows the details of the annotated dataset.
The sum of each category does not equal the total of reviews, as some reviews have been assigned to more than one label.

\subsubsection{Evaluation Metrics}
The performance of the classifiers is evaluated by \textit{precision}, \textit{recall}, and \textit{F1} as presented in related work \cite{Stanik2019,Maalej2016}.

\subsubsection{Experiments}
In this experiment, we compared the performance of ChatGPT and Guanaco-33B with ML models (Random Forest, Support Vector Machine), as well as various PTMs (BERT~\cite{Devlin2019}, CamemBERT\cite{Martin2020}, XLM-R~\cite{Ruder2019}), on the classification of app reviews.

\begin{figure}[!htb]
\centerline{
    \includegraphics[width=0.28\textwidth]{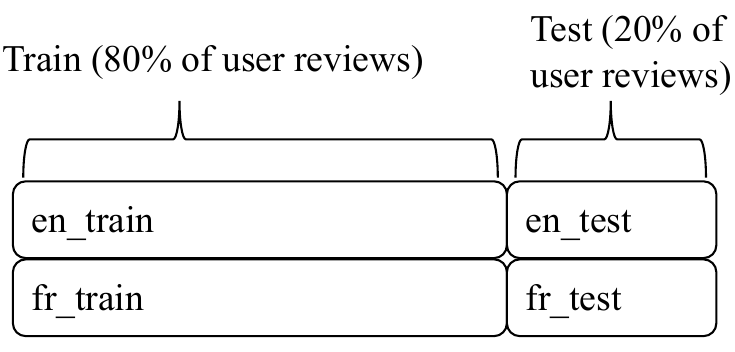}}
    \caption{Overview of dataset split for training and testing}
    \label{fig:dataset_split_lang}
\end{figure}

The ML models are trained using batch gradient descent, while PTMs employed mini-batch gradient descent, with a batch size of 12 and AdamW optimizer with a learning rate of $2e^{-5}$.
They are trained on 3 epochs on a machine with a NVIDIA Tesla T4 GPU with 16 GB VRAM.
The user reviews from the three apps of both languages were split using an 80:20 (training and test sets) ratio in a stratified manner, as illustrated in Figure \ref{fig:dataset_split_lang}.
We trained the classifiers using a combination of \textit{en\_train} and \textit{fr\_train}. 
Subsequently, the classifiers were individually tested on the \textit{en\_test} and \textit{fr\_test}.
We performed ten-fold cross-validation by randomly splitting the training and test sets ten times, and computed the average performance across these runs.

\begin{table}[!t]
\setlength{\tabcolsep}{3pt}
    \caption{Overview of the dataset for classification}
    \begin{tabular}{l | ccccc}
    \multicolumn{1}{c|}{App} & Language & Total & \multicolumn{1}{C{10mm}}{Feature request} & \multicolumn{1}{C{10mm}}{problem report} & Irrelevant \\
    \hline
    \rowcolor{lightgray}
    \cellcolor{white} & en & $2000$ & $223$ & $579$ & $1231$ \\
    \multirow{-2}{*}{Garmin Connect} & fr & $2000$ & $217$ & $772$ & $1051$ \\
    \hline
    \rowcolor{lightgray}
    \cellcolor{white} & en & $2000$ & $415$ & $876$ & $764$ \\
    \multirow{-2}{*}{Huawei Health} & fr & $2000$ & $387$ & $842$ & $817$ \\
    \hline
    \rowcolor{lightgray}
    \cellcolor{white} & en & $2000$ & $528$ & $500$ & $990$ \\
    \multirow{-2}{*}{Samsung Health}& fr & $2000$ & $496$ & $492$ & $1047$
    \end{tabular}
    \label{tab:classification-dataset}
\end{table}

We conduct classifications utilizing LLMs, specifically ChatGPT and Guanaco-33B, on all 12,000 user reviews.
This is executed under zero-shot setting, implying that no prior training is involved.
We have also assessed the performance of Guanaco-13B and Guanaco-65B.
However, the responses generated by Guanaco-13B are disorganized, thereby hindering the extraction of predicted labels using regular expressions.
The inference of Guanaco-65B is intolerably slow, even on advanced hardware such as the NVIDIA A100, making its usage impractical.

\begin{table*}[!htb]
\setlength{\tabcolsep}{5pt}
\caption{Classification accuracy on user reviews of three apps}
    \centering
    \begin{tabular}{l l | c c c | c c c | c c c | c c c}
    & \multirow{2}{*}{} &
    \multicolumn{3}{c|}{Feature Request} &
    \multicolumn{3}{c|}{Problem Report} &
    \multicolumn{3}{c|}{Irrelevant} &
    \multicolumn{3}{c}{Average Weight} \\
    & & $\mathit{Precision}$ & $\mathit{Recall}$  & $\mathit{F1}$  & $\mathit{Precision}$  & $\mathit{Recall}$  & $\mathit{F1}$  & $\mathit{Precision}$  & $\mathit{Recall}$  & $\mathit{F1}$  & $\mathit{Precision}$  & $\mathit{Recall}$  & $\mathit{F1}$  \\
    \hline
    \rowcolor{lightgray}\cellcolor{white} & Random Forest & 0.75 & 0.453 & 0.564 & 0.797 & 0.82 & 0.808 & 0.898 & 0.885 & 0.891 & 0.837 & 0.782 & 0.802 \\
    & SVM & \textbf{0.86} & 0.438 & 0.58 & 0.86 & 0.806 & 0.832 & 0.931 & 0.893 & 0.912 & 0.895 & 0.778 & 0.823 \\
    \rowcolor{lightgray}\cellcolor{white} & BERT & 0.814 & 0.782 & 0.797 & 0.897 & 0.914 & 0.905 & 0.972 & 0.954 & 0.963 & 0.918 & 0.909 & 0.913 \\
    & CamemBERT & 0.811 & 0.743 & 0.775 & 0.883 & 0.894 & 0.888 & 0.966 & 0.951 & 0.958 & 0.91 & 0.893 & 0.901 \\
    \rowcolor{lightgray}\cellcolor{white} &XLM-R & 0.823 & \textbf{0.811} & \textbf{0.816} & \textbf{0.902} & 0.917 & \textbf{0.909} & 0.979 & \textbf{0.958} & \textbf{0.968} & \textbf{0.925} & \textbf{0.917} & \textbf{0.92} \\
    & ChatGPT & 0.768 & 0.5 & 0.606 & 0.762 & \textbf{0.972} & 0.854 & \textbf{0.983} & 0.911 & 0.945 & 0.871 & 0.853 & 0.852 \\
    \rowcolor{lightgray}\multirow{-6}{*}{\cellcolor{white}En} & Guanaco-33B & 0.361 & 0.62 & 0.456 & 0.662 & 0.951 & 0.781 & 0.983 & 0.817 & 0.893 & 0.763 & 0.823 & 0.774 \\
    \hline 
    & Random Forest & 0.8 & 0.528 & 0.635 & 0.798 & 0.834 & 0.816 & 0.902 & 0.869 & 0.885 & 0.848 & 0.796 & 0.817 \\
    \rowcolor{lightgray}\cellcolor{white} &SVM & \textbf{0.895} & 0.459 & 0.606 & 0.86 & 0.828 & 0.844 & 0.956 & 0.89 & 0.922 & 0.912 & 0.791 & 0.838 \\
    & BERT & 0.766 & 0.725 & 0.744 & 0.871 & 0.866 & 0.869 & 0.947 & 0.931 & 0.939 & 0.888 & 0.872 & 0.88 \\
    \rowcolor{lightgray}\cellcolor{white} &CamemBERT & 0.852 & 0.823 & \textbf{0.837} & \textbf{0.922} & 0.925 & \textbf{0.923} & 0.977 & \textbf{0.96} & \textbf{0.968} & \textbf{0.936} & \textbf{0.924} & \textbf{0.929} \\
    & XLM-R & 0.819 & \textbf{0.833} & 0.825 & 0.917 & 0.921 & 0.919 & 0.982 & 0.949 & 0.965 & 0.93 & 0.919 & 0.924 \\
    \rowcolor{lightgray}\cellcolor{white} &ChatGPT & 0.853 & 0.473 & 0.608 & 0.782 & \textbf{0.973} & 0.868 & 0.978 & 0.935 & 0.956 & 0.888 & 0.866 & 0.863 \\
    \multirow{-6}{*}{\cellcolor{white}Fr}& Guanaco-33B & 0.296 & 0.576 & 0.391 & 0.624 & 0.97 & 0.759 & \textbf{0.985} & 0.756 & 0.856 & 0.737 & 0.797 & 0.739 \\
    \end{tabular}
    \label{tab:acc_classify}
\end{table*}

\subsubsection{Results}
The experiment results presented in Table \ref{tab:acc_classify} demonstrate that the ChatGPT exhibited good overall performance, which is comparable to that of ML models.
Its comparatively lower performance in classifying \textit{feature requests} can be attributed to the inherent complexity associated with such requests. 
In certain instances, users may express their desire for new features by criticizing existing ones or complaining about missing functionalities, rather than straightforwardly stating "I need...". 
Among all the models, XLM-R archived the best performance in bilingual classification.
Although the performance of ChatGPT does not match up to PTMs, it is noteworthy that these LLMs have achieved such performance without the utilization of any reviews during their training phase.
This suggests that ChatGPT can achieve satisfactory performance in user reviews of other application categories.

\subsection{Evaluation of Clustering}\label{sec:evaluation-clustering}
In this section, our objectives are to address the $RQ_2$: 
\textit{How semantically meaningful are the clusters generated by Mini-BAR?}

\subsubsection{Dataset}\label{sec:clustering-dataset}
In order to evaluate the performance of clustering algorithms, we created a dataset with 1200 user reviews.
We randomly selected 100 problem reports and 100 feature requests from each of the three apps in each of the two languages present in the dataset created in Section \ref{sec:classification-dataset}. 
Then the authors employed manual clustering for each collection of 200 bilingual reviews, all of which pertained to an identical category.
Reviews sharing the same topic were subsequently grouped into a single cluster.
In instances where a user's review mentioned multiple topics, the assignment of the cluster was determined by the initial topic that was reported.
Two authors independently performed the clustering of the 1200 reviews, and their individual results were later merged through discussion.
The resulting clusters were then considered as the ground truth for subsequent evaluation.

Table \ref{tab:clustering-dataset} shows the number of manually created clusters and the number of clusters whose size is greater than or equal to 5 in each category and app.
The feature requests encompass enhancements such as the modification of a Graphical User Interface (GUI), support for additional languages, increased activity options, integration with other applications, customization of permissions, and improvement of sleep tracking function.
The problem reports predominantly center on a series of issues, notably the application's unexpected crashes, errors encountered during the login process, difficulties in pairing with smartwatches, challenges with data synchronization, inconsistencies in the notification system, and complications related to Bluetooth connectivity, among others.

\begin{table}[!htb]
\setlength{\tabcolsep}{3pt}
\caption{Overview of manually created clusters}
    \begin{tabular}{l | C{1.2cm} C{1.2cm} C{1.2cm}}
    Bilingual & Garmin Connect & Huawei Health & Samsung Health \\
    \hline
    \rowcolor{lightgray} \#clusters in feature request & 89 & 74 & 69 \\
    \#clusters($size\geq5$) in feature request & 7 & 9 & 11 \\
    \rowcolor{lightgray} \#clusters in problem report & 45 & 44 & 41 \\
    \#clusters($size\geq5$) in problem report & 10 & 13 & 12 \\
    \end{tabular}
    \label{tab:clustering-dataset}
\end{table}

\subsubsection{Evaluation Metrics}
Following previous work~\cite{Wang2022}, we use two commonly used indices, \textit{Normalized Mutual Information (NMI)}~\cite{Vinh2009} and \textit{Adjusted Rand Index (ARI)}~\cite{Hubert1985}, to quantify the similarity between the automatic clustering and the ground truth.
\textit{NMI} ranges from 0 to 1, while \textit{ARI} ranges from -1 to 1.
A higher \textit{NMI} or \textit{ARI} indicates that the clustering method is more effective in producing clusters that align with the ground truth.
Note that the \textit{NMI} and \textit{ARI} are computed for clusters with a size of 5 or greater, as the minimum cluster size of HDBSCAN is set to 5.

\subsubsection{Experiments and Results}\label{sec:clustering-experiment}
In this section, we evaluate the performance of different text representation methods (including traditional frequency-based methods, \textit{bag of words (BOW)}, and \textit{TF-IDF}, as well as PTM-based methods, \textit{Universal Sentence Encoder}~\cite{Cer2018}, \textit{MiniLM}~\cite{Wang2020}, \textit{MPNet}~\cite{Song2020}, \textit{E5}~\cite{Wang2022E5} and \textit{Instructor}~\cite{Su2022}) on the dataset created in Section \ref{sec:clustering-dataset}. 
We performed three distinct experiments on English-only, French-only and bilingual user reviews. 
In each experiment, clustering is executed separately for the two categories (\textit{feature requests} and \textit{problem reports}) within each of the three applications.

\subsubsection{Results}
The average \textit{NMI} and \textit{ARI} of the three distinct experiments are shown in Table \ref{tab:clustering-eval}.
Results show that PTM-based methods outperformed traditional methods.
Among all text representation methods, \textit{Instructor} demonstrated the highest level of performance in clustering.
The results from the English user reviews clustering were superior to those derived from the bilingual and French user reviews clustering.
We attribute this variation to the relatively small French corpus employed for training the PTMs.

\begin{table}[t]
\centering
\caption{Evaluation on user reviews clustering}
\begin{tabular}{p{1.8cm} | ccc | ccc}
\multicolumn{1}{c|}{Embedding} & \multicolumn{3}{c|}{NMI} & \multicolumn{3}{c}{ARI} \\
\multicolumn{1}{c|}{Methods}      & en & fr & bi & en & fr & bi \\
\hline
\rowcolor{lightgray} BOW     &0.450&0.405&0.417&0.191&0.155&0.103\\
TF-IDF  &0.452&0.449&0.460&0.253&0.216&0.149\\
\rowcolor{lightgray} USE     &0.575&0.501&0.552&0.337&0.330&0.236\\
MiniLM  &0.548&0.567&0.541&0.323&\textbf{0.380}&0.219\\
\rowcolor{lightgray} MPNet   &0.616&0.575&0.593&0.400&0.346&0.278\\
E5      &0.465&0.401&0.436&0.248&0.190&0.139\\
\rowcolor{lightgray} Instructor &\textbf{0.713}&\textbf{0.587}&\textbf{0.603}&\textbf{0.597}&0.357&\textbf{0.308}
\\
\end{tabular}
\label{tab:clustering-eval}
\end{table}

While the proposed approach appears to have some degree of validity, the results do not appear to be particularly encouraging at this stage.
The primary reason for this is that our evaluation dataset is relatively small. 
Clustering algorithms require a sufficient amount of data to discover underlying patterns and structures. 
With a limited amount of data, it becomes difficult to identify meaningful groupings of text.
Moreover, text clustering is a challenging task, particularly given the informal nature of the terminology employed in user reviews and the prevalence of spelling errors. 
Additionally, based on our empirical analysis, it appears that longer user reviews tend to result in less accurate clustering.
This presents an opportunity for potential improvement by applying sentence-level clustering to user reviews.

\subsection{Evaluation of Summarization}\label{sec:evaluation-summarization}
In this section, we aim to investigate the $RQ_3$: \textit{: How effectively does ChatGPT perform in summarizing bilingual user reviews?}

\subsubsection{Dataset}
In Section \ref{sec:clustering-dataset}, we carried out manual clustering for a total of 1200 user reviews.
Among these, we utilized clusters with a size of 5 or greater to assess the performance of ChatGPT on summarization.

\subsubsection{Evaluation Metrics}
As outlined in Fabbri et al.~\cite{Fabbri2021}, human evaluators rate the generated summaries based on four dimensions: 
\textit{relevance} (the degree to which crucial information from the source has been included), 
\textit{consistency} (how well the summary aligns with the factual details of the source), 
\textit{fluency} (the quality of individual sentences), and 
\textit{coherence} (the overall quality and coherence of all the sentences in the summary).
Each dimension is scored on a Likert scale ranging from 1 to 5, with higher scores indicating superior performance.

\subsubsection{Experiments}
To evaluate the proficiency of ChatGPT in producing succinct English summaries of bilingual app reviews, we compare it with baseline approaches: 
Extractive summarization of Devine et al.~\cite{Devine2022}, which selects the most representative sentence from a cluster of app reviews as the summary, the sentence is chosen by calculating the similarity with all other sentences of that cluster.
Abstractive summarization with Guanaco models (13B, 33B and 65B version of Guanaco are used in our experiments)~\cite{Dettmers2023}. 
ChatGPT and Guanaco were instructed to synthesize clusters of user reviews into a single English sentence with the same prompt, as presented in Section \ref{sec:approach-summarization}.
Subsequently, human evaluators assessed the generated summaries.
Given the pairs of user reviews and corresponding summaries, two authors of this paper were asked to evaluate the summaries on the Likert scale in the four dimensions that were previously mentioned.

\begin{table}[!h]
\caption{Human evaluation on generated summaries}
\centering
\begin{tabular}{l | c c c c}
& Relevance & Consistency & Fluency & Coherence \\ \hline
\rowcolor{lightgray}Devine et al.~\cite{Devine2022} & 4.23 & 4.83 & 4.62 & 4.71 \\
Guanaco-13B & 3.67 & 3.68 & 4.92 & 4.91 \\
\rowcolor{lightgray}Guanaco-33B & 4.65 & 4.58 & 4.91 & 4.88 \\
Guanaco-65B & 4.79 & 4.77 & \textbf{4.95} & \textbf{4.94} \\
\rowcolor{lightgray}ChatGPT & \textbf{4.81} & \textbf{4.84} & \textbf{4.95} & \textbf{4.94}
\end{tabular}
\label{tab:summarization-result}
\end{table}

\subsubsection{Results} Table \ref{tab:summarization-result} presents the average results of hand-made evaluations.
Results show that abstractive approaches (ChatGPT, Guanaco-33B, and Guanaco-65B) generate very high quality sentences, and they are highly coherent when viewed in conjunction with one another.
On the other hand, the extractive approach excels at extracting the most significant information from a cluster; however, it falls short in capturing all the essential details.
During our manual evaluation, we found that Guanaco-13B tends to retrieve all available information available in the cluster without selectively focusing on crucial elements.
In contrast, Guanaco-33B performs significantly better in this regard by effectively filtering out non-essential information.
The results highlight the impressive performance of ChatGPT and Guanaco-65B in generating highly satisfactory summaries of user reviews.

\section{Threats to validity}
This section aims to identify potential threats to the validity of our study.

\subsubsection{App reviews from one category}
All reviews studied in this paper are collected from three Health \& Fitness apps (Garmin Connect, Huawei Health, and Samsung Health), mainly due to the context of health activity monitoring project. 
Instead of analyzing small number of reviews in many apps, we choose to annotate 12,000 reviews on three apps to create a larger dataset to evaluate clustering and summarization.
However, these three apps may not be representative of apps in other categories.
In the future, we will alleviate this threat by investigating user reviews of apps in various categories.

\subsubsection{Subjectivity in manual annotation}
The annotation of user reviews is a straightforward task, people without specific training can well classify or cluster the reviews.
However, subjectivity can still arise during manual annotation, leading to variations in how different annotators interpret and label the reviews.
To mitigate this threat, we 
(i) created a annotation guideline to detail the definition of each label following with examples, 
(ii) reviewed the final label through discussion and consensus.

\subsubsection{Issues of using ChatGPT}
ChatGPT has been selected as the classification and summarization component of Mini-BAR, owing to its superior capabilities.
However, it is noteworthy to mention that certain countries have imposed prohibitions on the use of ChatGPT. 
Some users may refrain from utilizing ChatGPT owing to concerns pertaining to data privacy.
And the cost of analyzing user reviews using ChatGPT cannot be overlooked, particularly in light of the large volume of user reviews.
To mitigate these challenges, we have implemented alternative strategies.
For classification, Mini-BAR users can utilize our XLM-R checkpoint, which has been trained on 12,000 reviews, or they can opt for Guanaco-33B. 
For summarization tasks, users have the option to use other large language models, specifically Guanaco-33B or Guanaco-65B.

\section{Conclusion}\label{sec:conclusion}
This paper introduces Mini-BAR, a mobile app review mining tool designed to assist app developers in extracting and summarizing user-reported issues and requests from a huge number of app reviews.
This tool is based on LLMs and operates under zero-shot setting.
Our empirical evaluation on the key steps of Mini-BAR resulted in numerous positive outcomes:
(i) it accurately classified bilingual user reviews with an F1 score of $0.85$;
(ii) it created meaningful clusters of bilingual user reviews with a \textit{NMI} greater than $0.6$;
(iii) it produced highly satisfactory summaries of bilingual user reviews.

In our future research, we intend to:
(i) conduct a comparative analysis of various prompts utilized for classification and summarization,
(ii) implement alternative large-scale language models for classification and summarization,
(iii) execute classification and clustering at the sentence level as opposed to the review level,
(iv) undertake evaluations using user reviews sourced from applications across a diverse range of categories.

\bibliographystyle{IEEEtran}
\bibliography{ref}

\end{document}